\documentclass[10pt,twocolumn,letterpaper]{article}

\usepackage[pagenumbers]{cvpr} 

\usepackage{tabularx,booktabs}
\usepackage{multirow}
\usepackage{algpseudocode}
\usepackage{algorithm}
\usepackage{float}
\usepackage[dvipsnames]{xcolor}
\usepackage{colortbl}
\usepackage{makecell}
\usepackage{gensymb}
\usepackage{amsmath}
\usepackage{graphicx}

\newif\ifdrafting
\draftingtrue 
\draftingfalse 
\ifdrafting
    \newcommand{\ds}[1]{{\color{red}[DS: #1]}}
    \newcommand{\cih}[1]{{\leavevmode\color[rgb]{0.2,0.6,0.2}[CH: #1]}}
    \newcommand{\df}[1]{{\color{blue}[DF: #1]}}
    \newcommand{\srbs}[1]{{\leavevmode\color[rgb]{1.0,0.4,0}[SS: #1]}}
    \newcommand{\jh}[1]{{\leavevmode\color[rgb]{0.8,0,0.8}[JH: #1]}}
    
\else
	\newcommand{\ds}[1]{}
	\newcommand{\cih}[1]{}
    \newcommand{\df}[1]{}
    \newcommand{\srbs}[1]{}
    \newcommand{\jh}[1]{}
	
\fi

\def\methodname{Diffusion for Metric Depth}

\def\ours{DMD}
\def\oursnk{\ours-NK}
\def\oursmix{\ours-MIX}

\definecolor{turquoise}{cmyk}{0.65,0,0.1,0.3}
\definecolor{purple}{rgb}{0.65,0,0.65}
\definecolor{dark_green}{rgb}{0, 0.5, 0}
\definecolor{orange}{rgb}{0.8, 0.6, 0.2}
\definecolor{red}{rgb}{0.8, 0.2, 0.2}
\definecolor{darkred}{rgb}{0.6, 0.1, 0.05}
\definecolor{blueish}{rgb}{0.0, 0.3, .6}
\definecolor{light_gray}{rgb}{0.7, 0.7, .7}
\definecolor{pink}{rgb}{1, 0, 1}
\definecolor{greyblue}{rgb}{0.25, 0.25, 1}

\definecolor{bestcol}{RGB}{254,196,79}
\newcommand{\best}[1]{\cellcolor{bestcol} \textbf{#1}}
\definecolor{secondbestcol}{RGB}{255,247,188}
\newcommand{\secondbest}[1]{\cellcolor{secondbestcol} #1}

\newcommand{\myparagraph}[1]{\smallskip\noindent\textbf{#1}}

\usepackage{amsthm,amsmath,amsfonts,bm,xspace}
\usepackage{bbm}
\usepackage{color}
\usepackage{enumitem}

\def\1{\bm{1}}

\def\vv{{\bm{v}}}

\def\veps{{\bm{\epsilon}}}

\def\vy{{\bm{y}}}

\DeclareMathAlphabet{\mathsfit}{\encodingdefault}{\sfdefault}{m}{sl}
\SetMathAlphabet{\mathsfit}{bold}{\encodingdefault}{\sfdefault}{bx}{n}

\newcommand{\E}{\mathbb{E}}

\makeatletter
\DeclareRobustCommand\onedot{\futurelet\@let@token\@onedot}
\def\@onedot{\ifx\@let@token.\else.\null\fi\xspace}

\makeatother

\definecolor{cvprblue}{rgb}{0.21,0.49,0.74}
\usepackage[pagebackref,breaklinks,colorlinks,citecolor=cvprblue]{hyperref}

\def\paperID{13181} 
\def\confName{CVPR}
\def\confYear{2024}

\title{Zero-Shot Metric Depth with a Field-of-View Conditioned Diffusion Model}

\author{Saurabh Saxena$^\dagger$, Junhwa Hur$^\ddagger$, Charles Herrmann$^\ddagger$, Deqing Sun$^\ddagger$, David J.\ Fleet$^\dagger$\thanks{DF is also affiliated with the University of Toronto and the Vector Institute.}\\
$^\dagger$Google DeepMind \ \ \ \ $^\ddagger$Google Research\\
{ \tt\small \{srbs,junhwahur,irwinherrmann,deqingsun,davidfleet\}@google.com}
}

\begin{document}
\maketitle
\begin{abstract}
\vspace*{-0.25cm}

While methods for monocular depth estimation have made significant strides on standard benchmarks, zero-shot metric depth estimation remains unsolved.  Challenges include the joint modeling of indoor and outdoor scenes, which often exhibit significantly different distributions of RGB and depth, and the depth-scale ambiguity due to unknown camera intrinsics.
Recent work \cite{bhat2023zoedepth} proposed a specialized multi-head architecture for jointly modeling indoor and outdoor scenes.
In contrast, we advocate a generic, task-agnostic diffusion model, with several advancements such as log-scale depth parameterization to enable joint modeling of indoor and outdoor scenes, conditioning on the field-of-view (FOV) to handle scale ambiguity and synthetically augmenting FOV during training to generalize beyond the limited camera intrinsics in training datasets.
Furthermore, by employing a more diverse training mixture than is common, and an efficient diffusion parameterization, our method, \methodname~(\ours) achieves a 25\% reduction in relative error (REL) on zero-shot indoor and 33\% reduction on zero-shot outdoor datasets over the current SOTA \cite{bhat2023zoedepth} using only a small number of denoising steps. For an overview see \href{https://diffusion-vision.github.io/dmd}{diffusion-vision.github.io/dmd}

\end{abstract}    
\vspace*{-0.1cm}
\section{Introduction}
\label{sec:intro}
\vspace*{-0.1cm}

Monocular estimation of metric depth in general environments, while useful for applications such as mobile robotics and autonomous driving, has proven elusive.
The two main barriers stem from (1) the large differences in RGB and depth distributions one finds in indoor and outdoor datasets, and (2) intrinsic scale ambiguity in images when one lacks knowledge of camera intrinsics.
Not surprisingly, most current models for monocular depth are either specific to indoor or outdoor scenes, or, if trained for both, they estimate scale-invariant depth. 

\begin{figure}[t]
    \centering
    \newcommand{\Figwidth}{1.0\linewidth}
    \includegraphics[width=\Figwidth]{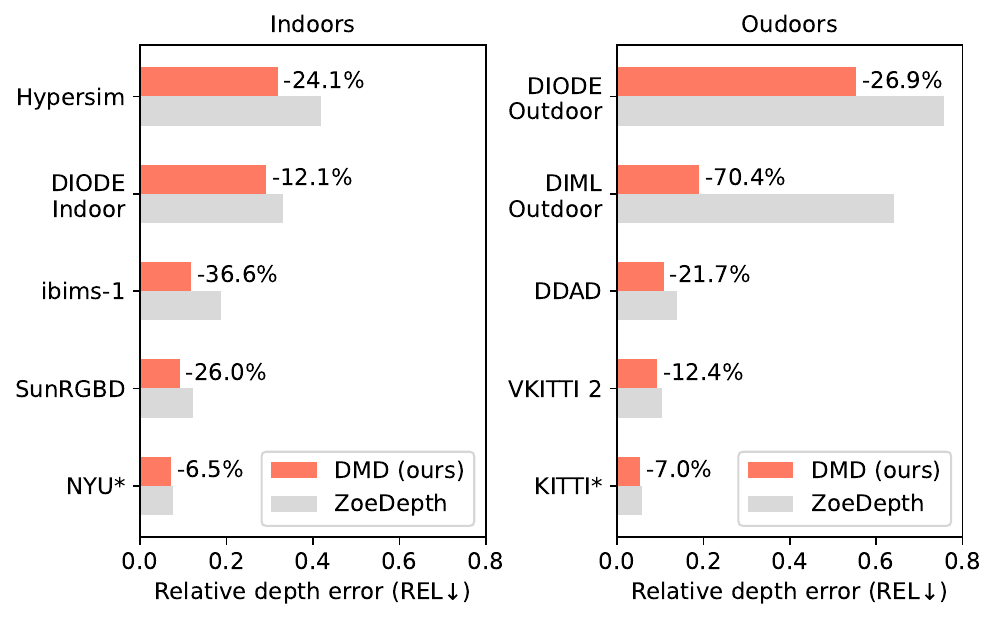}
    \vspace*{-0.55cm}
	\caption{Relative depth error 
	for \ours~(ours)  compared to ZoeDepth (SOTA) on eight zero-shot and two in-distribution ($\protect\ast$) benchmarks.
   \ours~outperforms ZoeDepth by a substantial margin on all benchmarks.
	}
    \label{fig:polar_plot}
    \vspace{-1em}
\end{figure}

Current models for metric depth are often trained solely on indoor or outdoor scenes, primarily on a single dataset captured with fixed camera intrinsics (e.g., with an RGBD camera, or, with RGB+LIDAR for outdoor scenes).
Such models avoid challenges caused by different indoor and outdoor depth distributions, but at the cost of generality. They also overfit to the camera intrinsics of the training dataset and do not generalize well to out of distribution data.

The predominant approach to jointly modeling indoor and outdoor data is to estimate scale- and shift-invariant depth, rather than metric depth (e.g., MiDaS \cite{midas}). By normalizing the depth distributions, one brings
indoor and outdoor depth distributions closer, and also avoids the problems of scale ambiguities in the presence of variable camera intrinsics.
Recently there has been growing interest in bridging these different approaches, training joint indoor-outdoor models that estimate metric depth. To cope with both indoor and outdoor domains, ZoeDepth \cite{bhat2023zoedepth} adds two heads to MiDaS \cite{midas}, one for each domain, to convert from scale-invariant depth to metric depth.

In this paper we advocate denoising diffusion models for zero-shot metric depth estimation, with the aid of several key innovations to obtain state of the art performance.
In particular, field-of-view (FOV) augmentation is used during training to improve generalization to different camera intrinsics, and FOV conditioning in training and inference helps to resolve intrinsic scale ambiguities, providing a further boost in performance. Representing depth in the log domain allocates model capacity in a more balanced way for indoor and outdoor scenes, improving indoor performance.
Finally, we find that $v$-parameterization in neural network denoising  greatly increases inference speed.
The resulting model, dubbed \ours~(\methodname), outperforms recently proposed metric depth model, ZoeDepth \cite{bhat2023zoedepth}.
In particular, \ours~yields much lower relative depth error than ZoeDepth on all eight out-of-distribution datasets in~\cite{bhat2023zoedepth} when fine-tuned on the same data. Expanding the training dataset leads to further improvement in performance (Fig.\  \ref{fig:polar_plot}).

\noindent
To summarize, we make the following contributions:
\begin{itemize}
    \item We introduce \ours, a simple yet effective method for zero-shot metric depth estimation on general scenes.
    \item We propose synthetically augmenting the FOV for improved generalization, FOV conditioning to resolve depth-scale ambiguity and representing depth in log-scale to better utilize the model's representation capacity.
    \item DMD establishes a SOTA on zero-shot metric depth,  achieving $25\%$ and $33\%$  lower relative error than  ZoeDepth on indoor and outdoor datasets, respectively, while being efficient, owing to the use of $v$-parameterization for diffusion. 
\end{itemize}

\section{Related Work}
\label{sec:related}

\myparagraph{Monocular depth (fine-tuned and evaluated in-domain).} 
Given the challenges of learning a joint indoor-outdoor model,
most approaches have restricted models to target a single dataset (either indoor or outdoor) with fixed intrinsics.
In this setting, great progress has been made with advancements in specialized architectures \cite{eigen2014depth,eigen2014predicting} such as the use of binning \cite{cao2016estimating,dorn,pixelformer,adabins,binsformer} or loss functions \cite{laina2016deeper,eigen2014predicting} that are suited for this task. \cite{MPSD_2020_ECCV} proposed combining multiple training datasets with variable intrinsics by normalizing the images to the same camera intrinsic.

\myparagraph{Joint indoor-outdoor models.}
To train joint indoor-outdoor models, one can mitigate the difficulty of learning diverse scene statistics by estimating scale- and shift-invariant depth instead. 
MiDaS \cite{midas} trains their model on diverse indoor-outdoor datasets and demonstrates good generalization to various unseen datasets. However, they do not provide metric depth.
DPT \cite{dpt} leverages this for pre-training and further fine-tunes separately for metric depth on NYU and KITTI.
ZoeDepth \cite{bhat2023zoedepth} proposes adding a mixture-of-experts head, supervised by scene type, on top of a similarly pre-trained model, thereby handling indoor and outdoor scenes.  
In contrast, our model, \ours, uses a relatively generic framework, without domain-specific architectural components.

\myparagraph{Intrinsics-conditioned monocular depth.}
Incorporating camera intrinsics for depth estimation has been briefly explored in previous work \cite{facil2019camconvs,he2018learning}.
They argue that intrinsic-conditioning allows one to train on multiple datasets with varying intrinsics,
but this is only demonstrated with small-scale experiments.
Similar to our method, ZeroDepth~\cite{guizilini2023towards} introduces an intrinsic-conditioned metric-scale depth estimator that is trained on large-scale training datasets.

While conditioning on the input field-of-view, we introduce a novel field-of-view augmentation scheme that augments training data by cropping or uncropping to simulate diverse FOVs and provide a large scale study on zero-shot generalization to both indoor and outdoor domains as well as robustness to diverse camera intrinsics.

\myparagraph{Diffusion for vision.}
Denoising diffusion models \cite{sohl2015deep,ho2020denoising} have recently emerged as a powerful class of generative models. 
Although initially proposed for natural image generation  \cite{sahariac-imagen,ho2021cascaded,nichol2021improved,dhariwal2021diffusion}, they have recently been shown to be effective for several computer vision tasks such as semantic segmentation \cite{ddp}, panoptic segmentation \cite{chen2023generalist}, optical flow \cite{ddvm} and monocular depth estimation \cite{ddp,diffusiondepth,ddvm}.
Ours is the first demonstration that diffusion models can also support state-of-the-art zero-shot metric depth estimation for general indoor or outdoor scenes.

\section{\methodname~(\ours)}
\label{sec:method}

In what follows we describe \ours~(\methodname) and  the design decisions to solve these issues.
In particular, we  cast monocular depth estimation as a generative RGB-to-depth translation task using denoising diffusion. 
To this end we introduce several technical innovations to conventional diffusion models and training procedures to accommodate zero-shot, metric depth.

\subsection{Diffusion models}
\label{sec:diffusion}
Diffusion models are probabilistic models that assume a forward process that gradually transforms a target distribution into a tractable noise distribution. A learned neural denoiser is trained to reverse this process, iteratively converting a noise sample to a sample from the target distribution.
They have been shown to be remarkably effective with images and video,
and they have recently begun to see use for dense vision tasks like segmentation, tracking, optical flow, and depth estimation.
They are attractive as they exhibit strong performance on regression tasks, capturing posterior uncertainty, without task specific architectures, loss functions and training procedures.

For \ours~we build on the task-agnostic Efficient U-Net architecture from DDVM~\cite{ddvm}. While DDVM used the $\epsilon$-parameterization for training the neural depth denoiser, here instead we use the $\vv$-parameterization \cite{salimans2022progressive}.
We find that the $\vv$-parameterization yields remarkably efficient inference, using as few as one or two refinement steps, without requiring 
progressive distillation \cite{salimans2022progressive}.

Under the $\vv$-parameterization, the denoising network is given a noisy target 
image (in our case a depth map), $\mathbf{z}_t = \alpha_t \mathbf{x} + \sigma_t \veps $, where $\mathbf{x}$ is the noiseless target input (depth map), 
$\veps  \sim \mathcal{N}(0, I)$, $t \sim \mathcal{U}(0, 1)$, $\sigma_t^2 = 1 - \alpha_t^2$, and $\alpha_t > 0$ is computed with a pre-determined noise schedule,
and the denoising network predicts $\vv \equiv \alpha_t\veps - \sigma_{t}\mathbf{x}$.
From the output of the denoising network, i.e., 
$\vv_{\theta} ( \mathbf{z}_t, \vy, t)$, where $\vy$ is an optional conditioning signal (RGB image in this case), one obtains an estimate of $\mathbf{x}$ at step $t$, i.e., $\hat{\mathbf{x}}_t = \alpha_t \mathbf{z}_t - \sigma_t \vv_{\theta} ( \mathbf{z}_t, \vy, t) $, and the corresponding estimate of the noise, denoted $\hat{\veps}_t$.
Under this parameterization, with a conventional L2 norm,
the training objective is based on the expected `truncated SNR weighting'  loss, i.e., 
 $\text{max}(\lVert \mathbf{x} - \hat{\mathbf{x}}_t \rVert_{2}^{2}, \lVert \veps - \hat{\veps}_t \rVert_{2}^{2})$ \cite{salimans2022progressive}.
Motivated by the superior performance of the L1 loss
in training DDVM \cite{ddvm} compared to the L2, we similarly employ a L1 loss for \ours~as well, yielding the  objective
\begin{equation}
    \E_{ \mathbf{x}, \vy, t, \veps} \left[
    \text{max}(\lVert \mathbf{x} - \hat{\mathbf{x}}_t \rVert_{1}, \lVert \veps - \hat{\veps}_t \rVert_{1}) \right] ~.
\label{eq:diffusion-loss}
\end{equation}

\begin{figure*}[t]
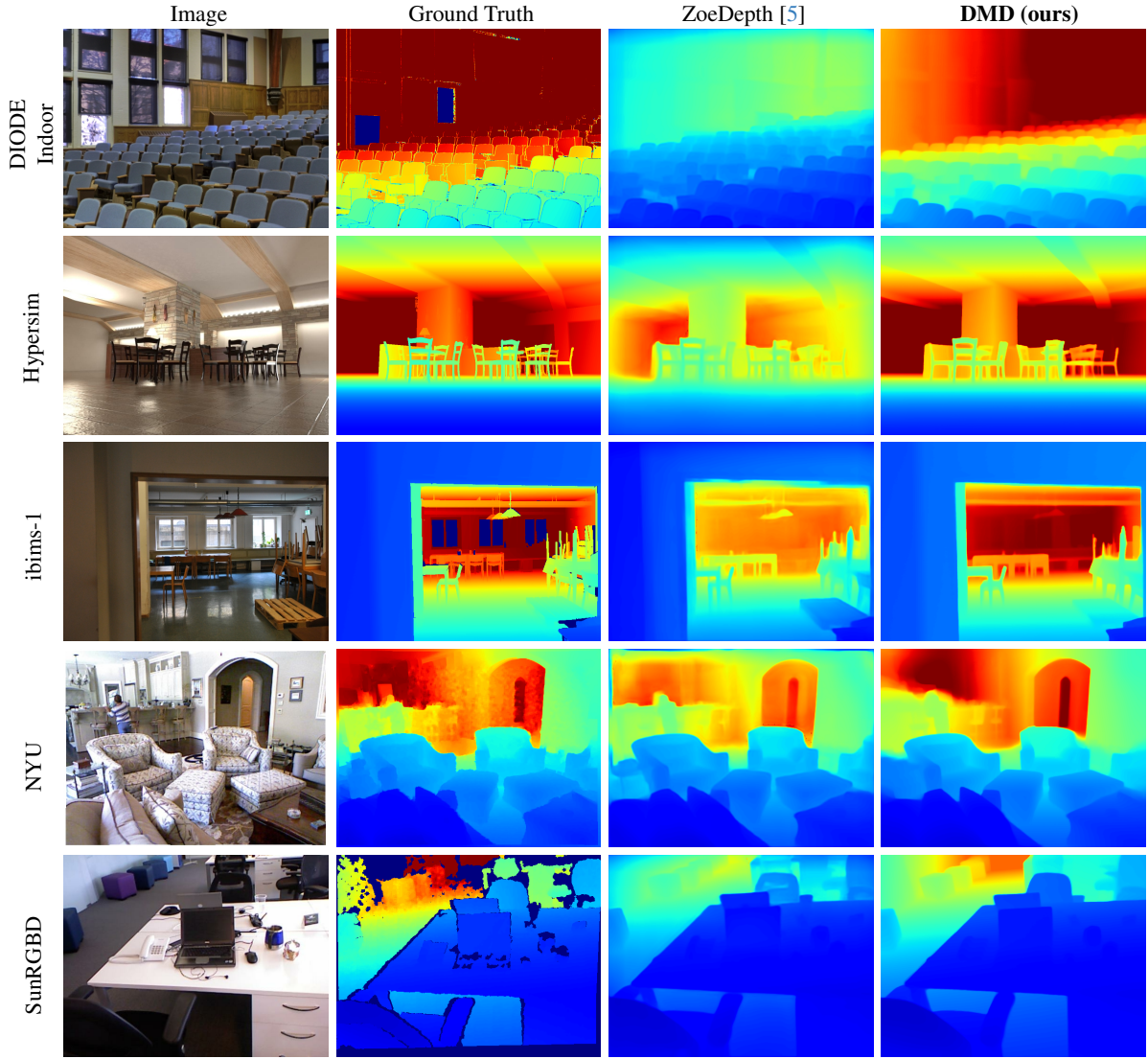

\setlength\tabcolsep{0.5pt}
\center
\small

\newcommand{\imgwidth}{0.21\textwidth}
\newcommand{\image}[3]{
\includegraphics[width=\imgwidth]{figures/qualitative_comparison/images/#1/#2_#3.png}
}
\newcommand{\imagejpg}[3]{
\includegraphics[width=\imgwidth]{figures/qualitative_comparison/images/#1/#2_#3.jpg}
}

\newcommand{\row}[2]{
\image{#1}{#2}{image} & \image{#1}{#2}{gt} & \image{#1}{#2}{depth_zd} & \image{#1}{#2}{depth_ours}
}
\newcommand{\rowrgbjpg}[2]{
\imagejpg{#1}{#2}{image} & \image{#1}{#2}{gt} & \image{#1}{#2}{depth_zd} & \image{#1}{#2}{depth_ours}
}

\begin{tabular}{ccccc}
    & Image & Ground Truth & ZoeDepth \cite{bhat2023zoedepth} & \textbf{\ours~(ours)} \\
    \rotatebox{90}{\hspace{2.3em} \makecell{DIODE\\Indoor}} &
    \rowrgbjpg{diode_indoor}{00017} \\
    \rotatebox{90}{\hspace{2em} Hypersim} &
    \rowrgbjpg{hypersim}{00700} \\
    \rotatebox{90}{\hspace{2.4em} ibims-1} & 
    \rowrgbjpg{ibims}{00081} \\
    \rotatebox{90}{\hspace{2.2em} NYU} & 
    \rowrgbjpg{nyu}{00190} \\
    \rotatebox{90}{\hspace{1.6em} SunRGBD} & 
    \rowrgbjpg{sunrgbd}{00658} \\
\end{tabular}
\vspace*{-0.2cm}
\caption{Qualitative comparison between our method and ZoeDepth \cite{bhat2023zoedepth} on indoor scenes.
Unlike ZoeDepth, our method estimates depths at more accurate scale over diverse datasets.}
\label{fig:ours_zd_samples_indoor}
\end{figure*}

\begin{table*}[!htb]
\footnotesize
\centering
\setlength{\tabcolsep}{3pt} 
\begin{tabular}
{@{}
    l@{\hspace{6pt}}
    c@{\hspace{6pt}}c@{\hspace{3pt}}c@{\hspace{3pt}}|
    c@{\hspace{6pt}}c@{\hspace{3pt}}c@{\hspace{3pt}}|
    c@{\hspace{6pt}}c@{\hspace{3pt}}c@{\hspace{3pt}}|
    c@{\hspace{6pt}}c@{\hspace{3pt}}c@{\hspace{3pt}}
    @{}}
\toprule
& \multicolumn{3}{c|}{\textbf{SUN RGB-D}} & \multicolumn{3}{c|}{\textbf{iBims-1 Benchmark}} & \multicolumn{3}{c|}{\textbf{DIODE Indoor}} & \multicolumn{3}{c}{\textbf{HyperSim}}\\

Method & $\delta_{1}$\,$\uparrow$ & REL\,$\downarrow$ & RMSE\,$\downarrow$ & $\delta_{1}$\,$\uparrow$ & REL\,$\downarrow$ & RMSE\,$\downarrow$ & $\delta_{1}$\,$\uparrow$ & REL\,$\downarrow$ & RMSE\,$\downarrow$ & $\delta_{1}$\,$\uparrow$ & REL\,$\downarrow$ & RMSE\,$\downarrow$   \\ 
\midrule
BTS~\cite{bts} & 0.740 & 0.172& 0.515 & 0.538 & 0.231& 0.919 & 0.210 & 0.418& 1.905 & 0.225 & 0.476& 6.404 \\
AdaBins~\cite{adabins}  & 0.771 & 0.159& 0.476  & 0.555 & 0.212& 0.901 & 0.174 & 0.443& 1.963  & 0.221 & 0.483& 6.546 \\
LocalBins~\cite{bhat2022localbins} & 0.777 & 0.156& 0.470 & 0.558  & 0.211 & 0.880 & 0.229  & 0.412 & 1.853 & 0.234  & 0.468 & 6.362 \\
NeWCRFs~\cite{yuan2022new} & 0.798 & 0.151& 0.424 & 0.548 & 0.206& 0.861 & 0.187 & 0.404& 1.867 & 0.255 & 0.442& 6.017 \\
\midrule
{ZoeD-M12-NK~\cite{bhat2023zoedepth}} & 0.856 &  0.123 & 0.356 & 0.615 & 0.186 & 0.777 & \secondbest{0.386} &  0.331 &   1.598 &   0.274 &    0.419 & 5.830 \\

\midrule
\textbf{\oursnk} & \secondbest{0.914} & \secondbest{0.109} & \secondbest{0.306} & \secondbest{0.801} & \secondbest{0.130} & \secondbest{0.563} & \best{0.402} & \secondbest{0.298} & \secondbest{1.407} & \secondbest{0.356} & \secondbest{0.382} & \secondbest{5.527} \\

\textbf{\oursmix} & \best{0.930} & \best{0.091} & \best{0.275} & \best{0.859} & \best{0.118} & \best{0.447} & 0.380 &  \best{0.291} & \best{1.292}  & \best{0.497}  & \best{0.318}  & \best{4.394} \\

\bottomrule
\end{tabular}
\vspace{-3pt}
\caption{\textbf{Zero-shot results} on unseen indoor datasets (evaluated at pixels with GT depth less 8m for SUN RGB-D, 10m for iBims-1 and DIODE Indoor, and 80m for Hypersim).
\colorbox{bestcol}{\textbf{Best}} and \colorbox{secondbestcol}{second-best} results are highlighted.
Trained on the same data (NYU and KITTI), \oursnk~outperforms ZoeD-M12-NK.
With more data (Taskonomy and nuScenes), \oursmix~outperforms by a much larger margin.
}
\label{tab:zero-shot-indoors}
\end{table*}

\begin{figure*}[t]
\setlength\tabcolsep{0.5pt}
\center
\small

\newcommand{\imgwidth}{0.22\textwidth}

\newcommand{\image}[3]{
\includegraphics[width=\imgwidth]{figures/qualitative_comparison/images/#1/#2_#3.png}
}

\newcommand{\imagejpg}[3]{
\includegraphics[width=\imgwidth]{figures/qualitative_comparison/images/#1/#2_#3.jpg}
}

\newcommand{\row}[2]{
\image{#1}{#2}{image} & \image{#1}{#2}{gt} & \image{#1}{#2}{depth_zd} & \image{#1}{#2}{depth_ours}
}

\newcommand{\rowrgbjpg}[2]{
\imagejpg{#1}{#2}{image} & \image{#1}{#2}{gt} & \image{#1}{#2}{depth_zd} & \image{#1}{#2}{depth_ours}
}

\begin{tabular}{ccccc}
    & Image & Ground Truth & ZoeDepth \cite{bhat2023zoedepth} & \textbf{\ours~(ours)} \\
    \rotatebox{90}{\hspace{1.7em} DDAD} &
    \rowrgbjpg{ddad}{02920} \\
    \rotatebox{90}{\hspace{1.3em} \makecell{DIML \\ Outdoor}} &
    \rowrgbjpg{diml_outdoor}{00292} \\
    \rotatebox{90}{\hspace{2.5em} \makecell{DIODE \\ Outdoor}} & 
    \rowrgbjpg{diode_outdoor}{00368} \\
    \rotatebox{90}{\hspace{1em} KITTI} & 
    \rowrgbjpg{kitti}{00166} \\
    \rotatebox{90}{\hspace{0.1em}  \makecell{Virtual \\ KITTI 2}} & 
    \rowrgbjpg{vkitti2}{00098} \\
\end{tabular}

\vspace*{-0.2cm}
\caption{Qualitative comparison between \ours~and ZoeDepth \cite{bhat2023zoedepth} on outdoor scenes. Compared with ZoeDepth \cite{bhat2023zoedepth}, our method is able to estimate a more accurate depth scale.}

\label{fig:ours_zd_samples_outdoor}
\end{figure*}

\begin{table*}[!htb]
\footnotesize
\centering
\setlength{\tabcolsep}{2.5pt} 
\begin{tabular}{@{}
    l@{\hspace{6pt}}
    c@{\hspace{6pt}}c@{\hspace{2.5pt}}c@{\hspace{2.5pt}}|
    c@{\hspace{6pt}}c@{\hspace{2.5pt}}c@{\hspace{2.5pt}}|
    c@{\hspace{6pt}}c@{\hspace{2.5pt}}c@{\hspace{2.5pt}}|
    c@{\hspace{6pt}}c@{\hspace{2.5pt}}c@{\hspace{2.5pt}}
    @{}}
\toprule
& \multicolumn{3}{c|}{\textbf{Virtual KITTI 2}} & \multicolumn{3}{c|}{\textbf{DDAD}} & \multicolumn{3}{c|}{\textbf{DIML Outdoor}} & \multicolumn{3}{c}{\textbf{DIODE Outdoor}} \\
Method & $\delta_{1}$\,$\uparrow$ & REL\,$\downarrow$ & RMSE $\downarrow$ & $\delta_{1}$\,$\uparrow$ & REL\,$\downarrow$ & RMSE $\downarrow$ & $\delta_{1}$\,$\uparrow$ & REL\,$\downarrow$ & RMSE $\downarrow$ & $\delta_{1}$\,$\uparrow$ & REL\,$\downarrow$ & RMSE $\downarrow$ \\ 
\midrule
BTS~\cite{bts} & 0.831 & 0.115 & 5.368 & 0.805 & 0.147 & 7.550 & 0.016 & 1.785 & 5.908 & 0.171 & 0.837 & 10.48 \\
AdaBins~\cite{adabins} & 0.826 & 0.122 & 5.420 & 0.766 & 0.154 & 8.560 & 0.013 & 1.941 & 6.272 & 0.161 & 0.863 & 10.35 \\

LocalBins~\cite{bhat2022localbins} & 0.810 & 0.127 & 5.981 & 0.777 & 0.151 & 8.139 & 0.016 & 1.820 & 6.706 & 0.170  &  {0.821} & 10.27 \\
NeWCRFs~\cite{yuan2022new} &  0.829 & 0.117 & 5.691 & \secondbest{0.874} & \secondbest{0.119} & \secondbest{6.183} & 0.010 & 1.918 & 6.283 & 0.176 & 0.854 & 9.228 \\
\midrule
{ZoeD-M12-NK~\cite{bhat2023zoedepth}}  &  0.850 & 0.105 & 5.095 &  0.824 & 0.138 & 7.225 & 0.292 & 0.641 & 3.610 & \best{0.208} & 0.757 & \best{7.569} \\
\midrule
\textbf{\oursnk}  & \secondbest{0.872}  &  \secondbest{0.093} & \secondbest{4.828} & 0.842 & 0.122 & 6.740 & \secondbest{0.544} & \secondbest{0.300} & \secondbest{2.522} & 0.162 & \secondbest{0.627} & 9.577  \\

\textbf{\oursmix} & \best{0.890} & \best{0.092}  & \best{4.387} & \best{0.907} & \best{0.108} & \best{5.365} & \best{0.602} & \best{0.190} & \best{2.089} & \secondbest{0.187} & \best{0.553} & \secondbest{8.943}  \\

\bottomrule
\end{tabular}
\vspace{-3pt}
\caption{\textbf{Zero-shot results} on four unseen outdoor datasets. \colorbox{bestcol}{\textbf{Best}} and \colorbox{secondbestcol}{second-best} results are highlighted.
Following pre-training, 
\oursnk~is trained on NYUv2 and KITTI, while \oursmix~adds Taxonomy and NuScenes for training. Using the same fine-tuning data, \oursnk~outperforms ZoeD-M12-NK on all benchmarks except DIODE Outdoor where it outperforms on REL but is behind on other metrics. \oursmix~shows further performance improvements over \oursnk~with the expanded training mixture.
}
\label{tab:zero-shot-outdoors}
\end{table*}

\subsection{Joint indoor-outdoor modelling}
\label{sec:joint-indoor-outdoor}

Training a joint indoor-outdoor model can be difficult because of the large differences in depth distributions one finds in indoor and outdoor scenes. Much of the available indoor training data have depths up to $10m$, while outdoor scenes include ground truth depths up to $80m$.
Further, training data is often lacking the variation in  camera intrinsics needed for robustness to images from different cameras.
Rather, many datasets are captured with a fixed camera.
To mitigate these issues we propose three innovations, namely, the use of log depth, field of view augmentation, and field of view conditioning.

\myparagraph{Log depth.}
\label{sec:log-scale}
Diffusion models usually model data distribution in [-1, 1].
One might convert metric depth to this range with linear scaling, i.e.,
\begin{equation}
d_\mathrm{lin} = \mathrm{normalize}(d_\mathrm{r}/d_\mathrm{max})~,
\end{equation}
where $d_\mathrm{r}$ is the raw depth in meters,
$d_\mathrm{max}$ is often taken to be 80m, to 
accommodate the usual outdoor depth range, and
$\mathrm{normalize}(d) = \mathrm{clip}(2*d - 1, -1, 1)$.
This, however, allocates little representation capacity to indoor scenes (where depths are usually less than $10m$).

Instead, we can allocate more representation capacity to indoor scenes with log-scaled depth ($d_\mathrm{log}$) as the target for inference, i.e.,
\begin{equation}
d_{\mathrm{log}} = \mathrm{normalize} \left( \frac{ \log(d_{\mathrm{r}}/d_{\mathrm{min}})}{\log(d_{\mathrm{max}}/d_{\mathrm{min}})} \right) ~,
\end{equation}
where $d_\mathrm{min}$ and $d_\mathrm{max}$ denote the minimum and maximum supported depths (e.g., $0.5m$ and $80m$). Empirically, we find log scaling to be remarkably beneficial.

\begin{figure*}[!h]
    \centering
    \small
    \setlength\tabcolsep{1pt} 
    \newcommand{\Figwidth}{0.245\linewidth}
    \newcommand{\trimleft}{0cm}
    \newcommand{\trimlower}{0cm}
    \newcommand{\trimright}{0cm}
    \newcommand{\trimupper}{0cm}
    \begin{tabular}{cccc}
    \includegraphics[trim={\trimleft{} \trimlower{}  \trimright{} \trimupper{}},clip,width=\Figwidth]{figures/metric_depth/rgb.jpg} &

        \includegraphics[trim={\trimleft{} \trimlower{}  \trimright{} \trimupper{}},clip,width=\Figwidth]{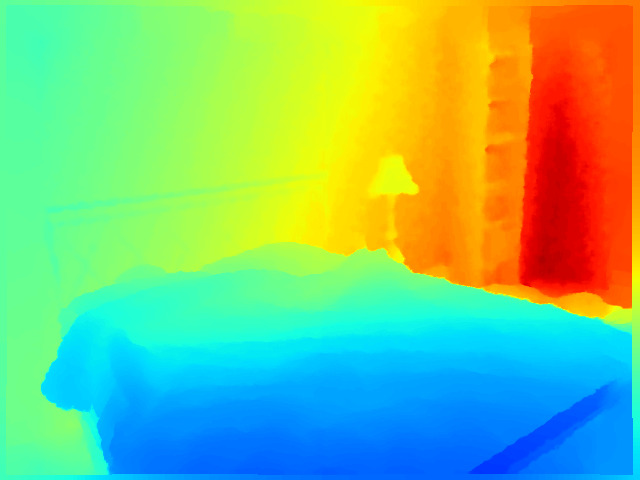} &
        
        \includegraphics[trim={\trimleft{} \trimlower{}  \trimright{} \trimupper{}},clip,width=\Figwidth]{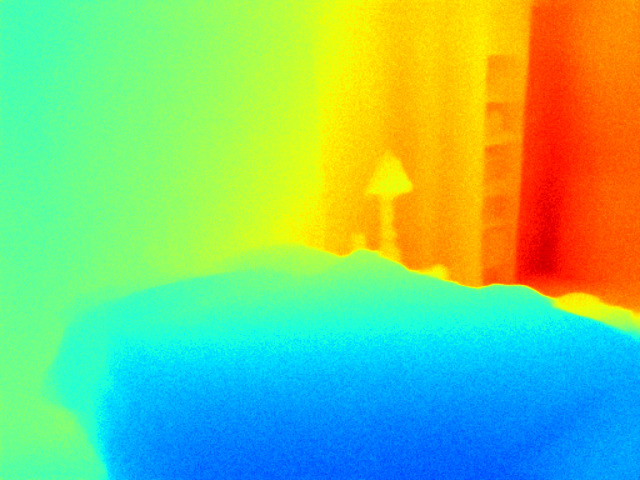} 
        &
        \includegraphics[trim={\trimleft{} \trimlower{}  \trimright{} \trimupper{}},clip,width=\Figwidth]{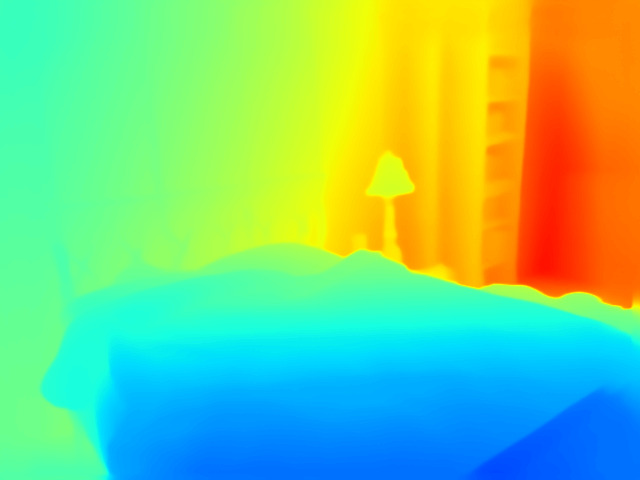}
        \\
        Image &
        Ground Truth &
        Prediction with linear-scaling
        & Prediction with log-scaling
        \\
        \end{tabular}
        \vspace*{-0.25cm}
	\caption{Linearly scaling depth leads to noisy predictions for images with shallow depth. See Section \ref{sec:joint-indoor-outdoor} for more details. Predicting depth in a log-scale fixes this. Note that here we use a max depth of 5 meters for better visualization.
	}
    \label{fig:metric_depth}
    \vspace*{-0.05cm}
\end{figure*}

\begin{table*}[!htb]
\footnotesize
\centering
\setlength{\tabcolsep}{2.5pt} 
\begin{tabular}{ll @{\hskip 3em} ccccc|ccccc}
\toprule
&& \multicolumn{5}{c|}{\textbf{Indoors}} & \multicolumn{5}{c}{\textbf{Outdoors}} \\
& Experiment & NYU & SunRGBD & DIODE Indoor & ibims-1 & Hypersim & KITTI & DIML Outdoor & DIODE Outdoor & Virtual KITTI 2 & DDAD \\
\midrule
\multirow{2}{*}{REL} & Linear scaling & 0.082 & \textbf{0.108} & 0.324 & 0.146 & 0.398 & 0.056 & 0.467 & 0.630 & \textbf{0.092} &  \textbf{0.122}  \\
& Log scaling &  \textbf{0.076} & 0.109 & \textbf{0.298} & \textbf{0.130} & \textbf{0.382} & \textbf{0.055} & \textbf{0.300} & \textbf{0.628} & 0.093 &    \textbf{0.122} \\
\midrule
\multirow{2}{*}{RMS} & Linear scaling &  0.340 & 0.314 & 1.526 & 0.612 & 5.693 & \textbf{2.516} & 3.126 & 10.129 & \textbf{4.788} & \textbf{6.288} \\
& Log scaling &  \textbf{0.313} & \textbf{0.306} & \textbf{1.407} & \textbf{0.563} & \textbf{5.527} & 2.527 & \textbf{2.522} & \textbf{9.577} & 4.828 & 6.740 \\
\bottomrule
\end{tabular}
\vspace{-3pt}
\caption{Ablation showing that log depth improves quantitative performance on indoor datasets which is understandable since log-scaling increases the share of representation capacity allocated to shallow depths.}
\label{tab:ablation-metric-depth}
\end{table*}

\myparagraph{Field-of-view augmentation.}
\label{sec:fov-aug}
Because datasets for depth estimation often have little or no variation in the field of view, it is easy for models to over-fit and thus generalize poorly to images with different camera intrinsics.
To encourage models to generalize well to different fields of view, we propose to augment
training data by cropping or uncropping to simulate diverse FOVs. While cropping is straightforward, for uncropping it is unclear how best to pad the enlarged image.  Our preliminary 
experiments used generative uncropping with Palette \cite{sahariac-palette}, however, we found that  padding the RGB image with Gaussian noise (mean-zero, variance 1) works as well, is simpler and more efficient.

For missing ground truth depth with uncropping augmentation, we adopt the approach in \cite{ddvm}, using a combination of near-neighbor in-filling and step-unrolled denoising during training.
It is shown in \cite{ddvm} that this technique is effective in coping with the inherent distribution shift between training and testing when ground truth data are noisy or incomplete.

\myparagraph{Field-of-view conditioning.}
Metric depth estimation from a single image is ill-posed in the presence of unknown camera intrinsics, as depth scales inversely with the field-of-view. While one might hope that diversifying camera intrinsics through FOV augmentation will help generalization to different cameras,  we and others \cite{wei2023fsdepth} observe that it is not sufficient in itself. FOV augmentations do help simulate  some variation in camera intrinsics, but variations in other factors, like focal length, are hard to simulate.

As a conditioning signal, to help disambiguate depth scale, we use $\tan(\theta/2)$, where $\theta$ is the vertical FOV. We explored conditioning on the horizontal FOV as well, but that did not improve results substantially.

\begin{figure*}[t]
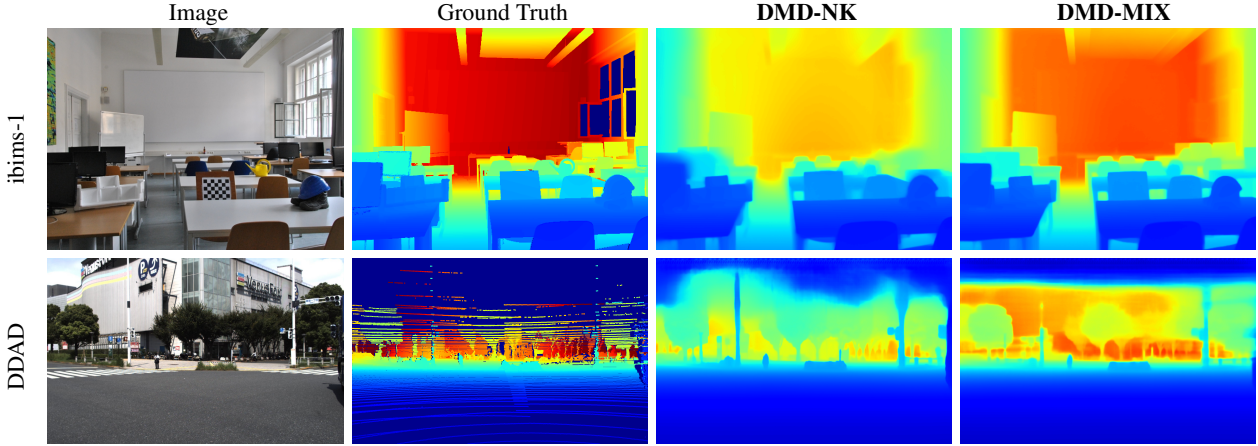

\setlength\tabcolsep{0.5pt}
\center
\small

\newcommand{\imgwidth}{0.225\textwidth}
\newcommand{\image}[3]{
\includegraphics[width=\imgwidth]{figures/qualitative_comparison/images_nk_ft/#1/#2_#3.png}
}
\newcommand{\imagejpg}[3]{
\includegraphics[width=\imgwidth]{figures/qualitative_comparison/images_nk_ft/#1/#2_#3.jpg}
}
\newcommand{\row}[2]{
\image{#1}{#2}{image} & \image{#1}{#2}{gt} & \image{#1}{#2}{depth_ours_nk} & \image{#1}{#2}{depth_ours_ft}
}
\newcommand{\rowrgbjpg}[2]{
\imagejpg{#1}{#2}{image} & \image{#1}{#2}{gt} & \image{#1}{#2}{depth_ours_nk} & \image{#1}{#2}{depth_ours_ft}
}

\begin{tabular}{c @{\hskip 1em} cccc}
    & Image & Ground Truth & \textbf{\oursnk} & \textbf{\oursmix} \\
    \rotatebox{90}{\hspace{2.5em} ibims-1} & 
    \rowrgbjpg{ibims}{00036} \\
    \rotatebox{90}{\hspace{2em} DDAD} &
    \rowrgbjpg{ddad}{00050} \\
\end{tabular}
\vspace*{-0.2cm}
\caption{Qualitative comparison between \oursnk~(fine-tuned on NYU and KITTI) and \oursmix~(fine-tuned on KITTI, NYU, nuScenes, and Taskonomy). \oursmix~further improves depth scale estimation as well as fine details on depth boundaries.}
\label{fig:ours_nk_mix}
\end{figure*}

\section{Experiments}
\label{sec:experiments}


\begin{table*}[t]
\centering
\footnotesize
\setlength{\tabcolsep}{5.5pt}
\begin{tabularx}{\linewidth}{lccccccccccccc}
\toprule
\multirow{2}{*}{\textbf{Method}} & \multicolumn{6}{c}{NYU} & \multicolumn{7}{c}{KITTI} \\
\cmidrule(r){2-7} \cmidrule(l){8-14}
& \textbf{$\delta_1$}$\uparrow$       & \textbf{$\delta_2$}$\uparrow$  & \textbf{$\delta_3$}~$\uparrow$            & REL~$\downarrow$          & RMS~$\downarrow$  & $\log_{10}$~$\downarrow$ & \textbf{$\delta_1$}$\uparrow$ & \textbf{$\delta_2$}$\uparrow$ & \textbf{$\delta_3$}$\uparrow$ & REL~$\downarrow$ & Sq-rel~$\downarrow$ & RMS~$\downarrow$ & RMS log~$\downarrow$ \\
\midrule
\multicolumn{14}{@{}p{\textwidth}}{\textit{Domain-specific models:}\hfill} \\
BTS~\cite{bts}& {0.885}          & 0.978          & 0.994          & 0.110            & {0.392}          & {0.047} & 0.956 & 0.993 & {0.998} & 0.059   & 0.245  & 2.756 & 0.096  \\ 
{DPT \citep{dpt}} & 0.904 & 0.988 & {0.998} & 0.110 & 0.357 & 0.045  & 0.959 & {0.995} & {0.999} & 0.062 & -- & 2.573 & 0.092 \\

AdaBins~\cite{adabins} & {0.903} & {0.984} & {0.997} & {0.103}     & {0.364} & {0.044}  & 0.964 & {0.995} & {0.999} & 0.058  & 0.190  & 2.360 & 0.088 \\ 
{NeWCRFs~\cite{yuan2022new}}  & 0.922 & 0.992 & {0.998} & {0.095} & {0.334} & {0.041} &  0.974 &  {0.997} &  {0.999} & 0.052 & 0.155 & 2.129 & 0.079 \\ 

{BinsFormer \citep{binsformer}} & 0.925 & 0.989 & 0.997 & 0.094 & 0.330 & 0.040  & {0.974} & {0.997} & {0.999} & {0.052} & {0.151} & {2.098} & {0.079} \\
{PixelFormer \citep{pixelformer}} & 0.929 & 0.991 & {0.998} & 0.090 & 0.322 & 0.039  & {0.976} & {0.997}  & {0.999} & {0.051} & {0.149} & {2.081} & {0.077} \\
IEBins \cite{shao2023IEBins} & 0.936 &  0.992 &  {0.998} & 0.087  &  0.314 & 0.038  & {0.978}  & {0.998}  & {0.999} & {0.050}  &  {0.142}  &  {2.011}  & {0.075} \\
{MIM \citep{xie2022revealing}} & {0.949} & {0.994} & {0.999} & 0.083 & {0.287} & 0.035 & {0.977} & {0.998} & {1.000} & {0.050} & {0.139} & {1.966} & {0.075} \\
DDVM \cite{ddvm} & 0.946 & 0.987 & 0.996 & {0.074} & 0.315 & {0.032} &  0.965 &  0.994 &  {0.998} & 0.055 & 0.292 & 2.613 & 0.089\\
\midrule
\multicolumn{14}{@{}p{\textwidth}}{\textit{Joint indoor-outdoor models:}\hfill} \\
ZoeD-M12-NK  & \textbf{0.953} & \textbf{0.995} & \textbf{0.999} & 0.077 & \textbf{0.277} & 0.033  &  0.966 &   0.993 &   0.996 &   0.057 &   0.204 &   \textbf{2.362} &   0.087 \\ 
\textbf{\oursnk}  & 0.944 & 0.986 & 0.996 & 0.076 & 0.313 & 0.033 & 0.964 & 0.994 & \textbf{0.999} & 0.055 & 0.219 & 2.527 & 0.087 \\
\textbf{\oursmix}  & \textbf{0.953} & 0.989 & 0.996 & \textbf{0.072} & 0.296 & \textbf{0.031} & \textbf{0.967}  & \textbf{0.995} & \textbf{0.999} & \textbf{0.053} & \textbf{0.203}  & 2.411 & \textbf{0.084} \\

\bottomrule
\end{tabularx}
\caption{\textbf{In-domain results} showing better relative error than ZoeDepth for our models on both the NYU and KITTI datasets. \textbf{Best} results (amongst indoor-outdoor models only) are bolded. For reference, we provide results for models trained separately for the indoor and outdoor domains showing that our results are competitive despite being a more general model.}
\label{tab:results-nyu-kitti}
\end{table*}

\subsection{Training data}
\label{sec:training-data}

DDVM \cite{ddvm} showed that using large amounts of diverse training data is important for generic  models with no task-specific inductive biases. Here we follow the training strategy proposed in \cite{ddvm}, initializing the model with unsupervised pre-training on ImageNet \cite{imagenet_cvpr09} and Places365 \cite{zhou2017places}, with tasks proposed in \cite{sahariac-palette}.
This is followed by supervised pre-training on 
ScanNet \cite{scannet}, SceneNet-RGBD  \cite{scenenet-rgbd}, and Waymo  \cite{waymoOpenDataset}. Unlike \cite{ddvm}, we also include the DIML Indoor \cite{cho2021diml} dataset for more diversity.  No FOV augmentation or conditioning is used for this pre-training stage.

For the final training stage we train on a mixture of NYU \cite{nyu}, Taskonomy \cite{taskonomy2018}, KITTI \cite{kitti} and nuScenes \cite{nuscenes}.
At this stage we apply FOV augmentation to NYU, KITTI and nuScenes, but not Taskonomy as it is large and has substantial FOV diversity and also add the FOV conditioning.
Finally, to enable fair comparisons with the current SOTA ZoeDepth \cite{bhat2023zoedepth} model, we also train a model solely from NYU and KITTI, like ZoeDepth.

\begin{table*}[!htb]
\footnotesize
\centering
\setlength{\tabcolsep}{2.5pt} 
\begin{tabular}{ll@{\hskip 3em} ccccc|ccccc}
\toprule
&& \multicolumn{5}{c|}{\textbf{Indoors}} & \multicolumn{5}{c}{\textbf{Outdoors}} \\
Metric & Experiment & NYU & SunRGBD & DIODE Indoor & ibims-1 & Hypersim & KITTI & DIML Outdoor & DIODE Outdoor & Virtual KITTI 2 & DDAD \\
\midrule
\multirow{2}{*}{REL} & No FOV cond &   0.081 & 0.116 & 0.316 & 0.18 & 0.400 & 0.057 & 1.257 & \textbf{0.613} & 0.100 &  \textbf{0.121} \\
& With FOV cond &  \textbf{0.076} & \textbf{0.109} & \textbf{0.298} & \textbf{0.130} & \textbf{0.382} & \textbf{0.055} & \textbf{0.300} & 0.628 & \textbf{0.093} &  0.122 \\
\midrule
\multirow{2}{*}{RMS} & No FOV cond & 0.319 & 0.325 & 1.474 & 0.712 & \textbf{5.196} & 2.574 & 5.382 & \textbf{8.582} & 5.021 & 6.826\\
& With FOV cond & \textbf{0.313} & \textbf{0.306} & \textbf{1.407} & \textbf{0.563} & 5.527 & \textbf{2.527} & \textbf{2.522} & 9.577 & \textbf{4.828} & \textbf{6.740} \\
\bottomrule
\end{tabular}
\caption{Depth errors for models trained with and without field-of-view conditioning. 
Results shows that FOV conditioning provides a substantial boost in performance.
DIML Outdoor benefits the most, which is understandable given its large FOV, for which generalization is a major challenge for simple FOV augmentation.}
\label{tab:ablation-fov}
\end{table*}

\begin{figure}[t]
   \vspace*{-0.1cm}
    \centering
    \scriptsize
    \setlength\tabcolsep{1pt} 
    \newcommand{\Figwidth}{0.46\linewidth}
    \begin{tabular}{ccc}
      & \hspace{2em}\textbf{Indoors} & \hspace{1.5em}\textbf{Outdoors} \\
    \rotatebox{90}{\hspace{2em}\textbf{In-distribution}} & \includegraphics[width=\Figwidth]{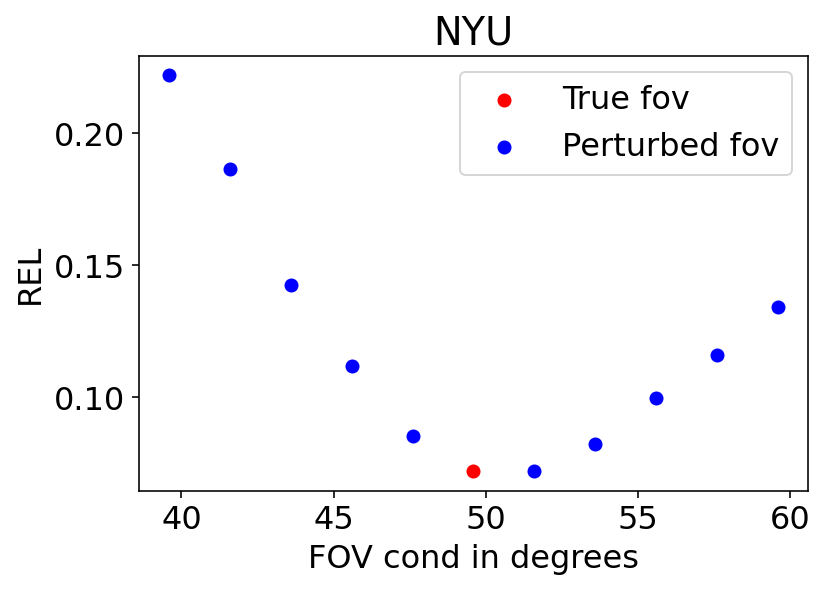} &
        \includegraphics[width=\Figwidth]{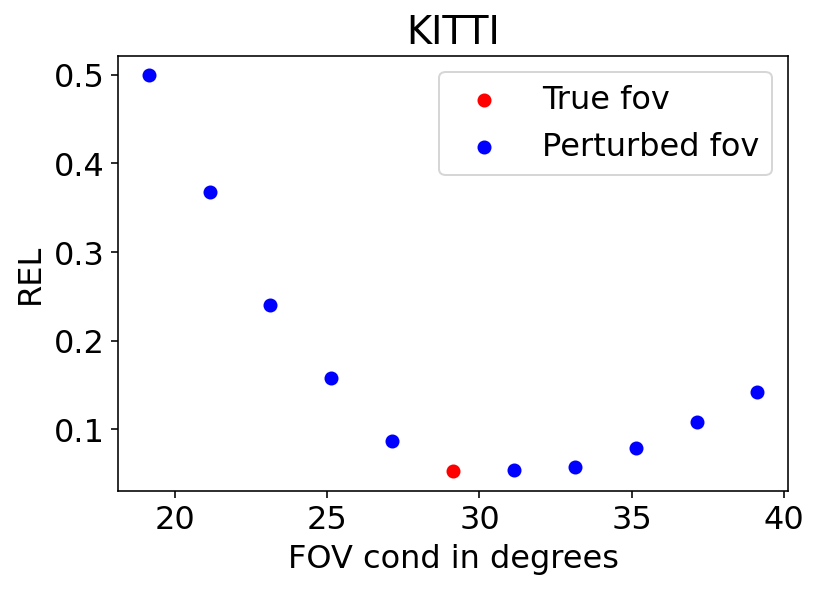} \\
   \rotatebox{90}{\hspace{4.5em}\textbf{OOD}} & \includegraphics[width=\Figwidth]{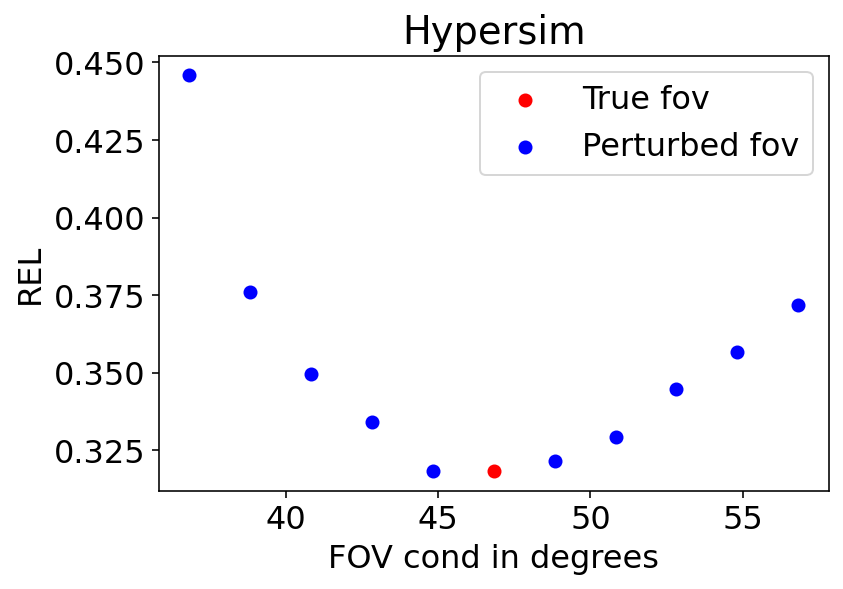} &
    \includegraphics[width=\Figwidth]{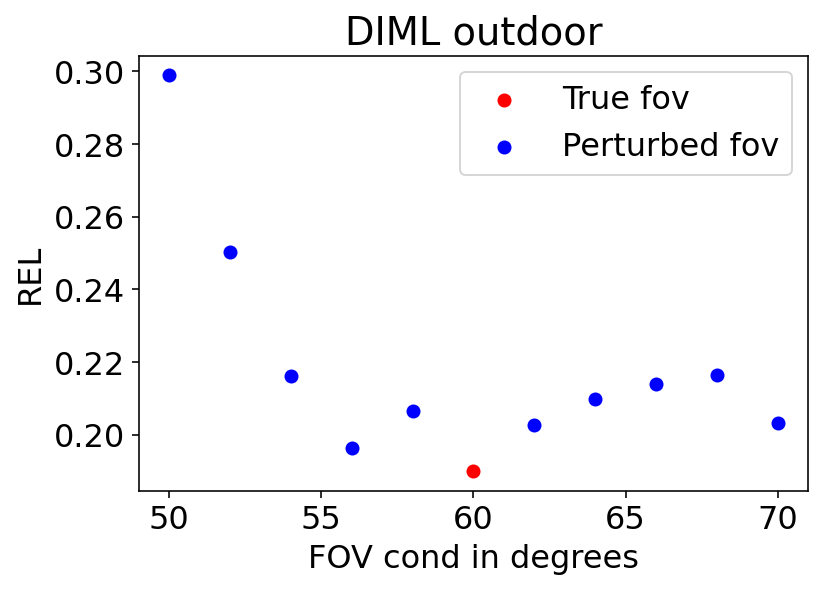} \\
        \end{tabular}
        \vspace*{-0.3cm}
	\caption{Plots showing the effect of perturbing the FOV during inference. Optimal performance is at or near the true FOV.  Performance degrades with larger perturbation.
	}
    \label{fig:fov_perturb}
    \vspace*{-0.25cm}
\end{figure}

 
\subsection{Design choices}
\label{sec:design_choices}

\myparagraph{Denoiser Architecture.}
We adopt the modifications of the \textit{Efficient U-Net} \cite{sahariac-imagen} proposed in DDVM \cite{ddvm}, with one further modification to support FOV conditioning. The FOV embedding, like the timestep embedding, is constructed by first building a sin-cos positional embedding \cite{vaswani2017attention} followed by linear projection. The sum of these two embeddings is used to modulate different layers of the denoiser backbone using FiLM \cite{perez2018film} layers. Following \cite{bhat2023zoedepth} we train at a resolution of 384$\times$512. The predicted depth maps are resized to the ground-truth resolution for evaluation, following prior work \cite{bhat2023zoedepth}. Other training hyper-parameters such as the batch size and optimizer details are like those in  \cite{ddvm}.

\myparagraph{Augmentations.}
In addition to the FOV augmentation (Sec.\  \ref{sec:fov-aug}) we use random horizontal flip augmentation, like many prior works.

\myparagraph{Sampler.}
We use the DDPM \cite{ho2020denoising} sampler with eight denoising steps for indoor datasets. For outdoor datasets we find that two denoising steps suffice.
We report results using a mean of eight samples, following \cite{ddvm}; as  shown in Table \ref{tab:ablation-multiple-samples}, sample averaging leads to improved performance.

\vspace*{0.1cm}
\myparagraph{Evaluation.} We adopt the evaluation protocol of ZoeDepth \cite{bhat2023zoedepth}. We report in-distribution performance on the NYU \cite{nyu} and KITTI \cite{kitti} datasets, and generalization performance on eight unseen datasets \cite{bhat2023zoedepth}, namely, SunRGBD \cite{sunrgbd}, iBims-1 \cite{sunrgbd}, DIODE Indoor \cite{diode_dataset}, Hypersim \cite{hypersim} for indoors, and Virtual KITTI 2 \cite{cabon2020vkitti2}, DDAD \cite{ddad}, DIML Outdoor \cite{cho2021diml}, DIODE Outdoor \cite{diode_dataset} for outdoors. We closely follow the evaluation protocol, including depth ranges and cropping, used in \cite{bhat2023zoedepth} and report results using the standard error and accuracy metrics that are used in literature.

\begin{table*}[!htb]
\footnotesize
\centering
\setlength{\tabcolsep}{2.5pt} 
\begin{tabular}{lcccccc|ccccc}
\toprule
&& \multicolumn{5}{c|}{\textbf{Indoors}} & \multicolumn{5}{c}{\textbf{Outdoors}} \\
Metric &\# samples & NYU & SunRGBD & DIODE Indoor & ibims-1 & Hypersim & KITTI & DIML Outdoor & DIODE Outdoor & Virtual KITTI 2 & DDAD \\
\midrule
\multirow{2}{*}{REL} & 1 & 0.077 & 0.110 & 0.304 & 0.135 & 0.383 & \textbf{0.055} & 0.311 & 0.634 & \textbf{0.093} & \textbf{0.122} \\
& 8 & \textbf{0.076} & \textbf{0.109} & \textbf{0.298} & \textbf{0.130} & \textbf{0.382} & \textbf{0.055} & \textbf{0.300} & \textbf{0.628} & \textbf{0.093} & \textbf{0.122} \\
\midrule
\multirow{2}{*}{RMS} & 1 & 0.321 & 0.311 & 1.426 & 0.579 & \textbf{5.521} & 2.544 & 2.572 & 9.602 & 4.842 & 6.755 \\
& 8 & \textbf{0.313} & \textbf{0.306} & \textbf{1.407} & \textbf{0.563} & 5.527 & \textbf{2.527} & \textbf{2.522} & \textbf{9.577} & \textbf{4.828} & \textbf{6.740} \\
\bottomrule
\end{tabular}
\caption{Averaging multiple samples leads to small but consistent improvement on both REL and RMS.}
\label{tab:ablation-multiple-samples}
\end{table*}
\subsection{Results}
\label{sec:results}

\noindent
\textbf{Zero-shot.} Tables \ref{tab:zero-shot-indoors} and \ref{tab:zero-shot-outdoors} report \textit{zero-shot} performance on eight OOD datasets.  \oursnk~permits fair comparison with ZoeDepth, as both are fine-tuned on NYU and KITTI.
With these datasets, \ours~outperforms ZoeDepth across all but DIODE Outdoor, where \oursnk~outperforms ZoeDepth on relative error but not RMSE and $\delta_1$.
Importantly, but perhaps not surprisingly, further performance improvements are obtained by fine-tuning on larger amounts of data.
Our variant trained on a mixture with KITTI, NYU, nuScenes, and Taskonomy, dubbed \oursmix,  generalizes much better on OOD test data, establishing an even stronger new state of the art.

Figures \ref{fig:ours_zd_samples_indoor} and \ref{fig:ours_zd_samples_outdoor} visualize qualitative comparison between \ours~and ZoeDepth on indoor and outdoor datasets respectively. 
Our method captures more accurate metric-scale depth with both indoor and outdoor scenes.
\cref{fig:ours_nk_mix} illustrates the qualitative differences in depth maps from \oursnk~and \oursmix.
By virtue of training on a larger dataset, \oursmix~significantly improves the depth scale and  fine-grained depth details near object boundaries.

\vspace*{0.2cm}
\noindent
\textbf{In-distribution.} Table \ref{tab:results-nyu-kitti} reports results on the KITTI and NYU datasets.
On KITTI, \oursmix~outperforms ZoeDepth on all metrics except RMSE where it is competitive. 
On NYU \oursmix~substantially outperforms ZoeDepth on relative error and is competitive on other metrics.
Interestingly, \oursmix~outperforms \oursnk~on in-distribution data.
\begin{table*}[!htb]
\footnotesize
\centering
\setlength{\tabcolsep}{2.5pt} 
\begin{tabular}{llccccc|ccccc}
\toprule
&& \multicolumn{5}{c|}{\textbf{Indoors}} & \multicolumn{5}{c}{\textbf{Outdoors}} \\
Metric & Experiment & NYU & SunRGBD & DIODE Indoor & ibims-1 & Hypersim & KITTI & DIML Outdoor & DIODE Outdoor & Virtual KITTI 2 & DDAD \\
\midrule

\multirow{2}{*}{REL} & No FOV aug or cond & \textbf{0.074} & 0.124 & 0.337 & 0.180 & 0.479 & \textbf{0.055} & 1.399 & \textbf{0.615} & 0.095 & \textbf{0.116} \\
& W/ FOV aug and cond &  0.076 & \textbf{0.109} & \textbf{0.298} & \textbf{0.130} & \textbf{0.382} & \textbf{0.055} & \textbf{0.300} & 0.628 & \textbf{0.093} & {0.122} \\
\multirow{2}{*}{RMS} & No FOV aug or cond &  \textbf{0.310} & 0.348 & 1.535 & 0.722 & \textbf{5.247} & 2.597 & 5.919 & \textbf{8.529} & 4.874 & \textbf{6.476} \\
& W/ FOV aug and cond &  0.313 & \textbf{0.306} & \textbf{1.407} & \textbf{0.563} & 5.527 & \textbf{2.527} & \textbf{2.522} & 9.577 & \textbf{4.828} & 6.740 \\
\bottomrule
\end{tabular}
\caption{Ablation showing that training without FOV augmentations and conditioning hurts generalization to out-of-domain data due to overfitting on the training data intrinsics.}
\label{tab:ablation-fov-aug-cond}
\end{table*}

\subsection{Ablations}
\label{sec:ablations}

We next consider several ablations to test different components of the model, all with \oursnk~for expedience.


\vspace*{0.2cm}
\noindent
\textbf{Log vs linearly scaled depth.} 
Table \ref{tab:ablation-metric-depth} shows that parameterizing depth in log scale (Sec.\ \ref{sec:log-scale}) improves quantitative performance. As expected, this is beneficial for datasets of indoor scenes and also for datasets of outdoor scenes with shallower depths, like DIML Outdoor and DIODE Outdoor.
Furthermore, as shown in Fig.\ \ref{fig:metric_depth}, using linear-scaling leads to noise artifacts in the depth estimates for indoor scenes which is resolved with log-depth parameterization.

\vspace*{0.2cm}
\noindent
\textbf{Field-of-view conditioning.} 
Table \ref{tab:ablation-fov} shows that FOV conditioning achieves the best performance.
Fig.\ \ref{fig:fov_perturb} perturbs the conditioning FOV signal during inference, showing that optimal performance occurs at or close to the true FOV.

\begin{table}[t]
\footnotesize
\setlength{\tabcolsep}{3pt} 
\centering
\begin{tabular}{c|cc|cc}
\toprule
                  & \multicolumn{2}{c|}{\textbf{REL $\downarrow$}} & \multicolumn{2}{c}{\textbf{RMSE $\downarrow$}} \\

 & {NYU}   & {KITTI} & {NYU}   & {KITTI} \\ \midrule
ZoeDepth Auto Router & 0.102 & 0.075 & 0.377 & \textbf{2.584} \\
Ours no fov aug / cond & \textbf{0.074} & \textbf{0.055} &  \textbf{0.310} & 2.597   \\
\bottomrule
\end{tabular}
\caption{Comparison against ZoeDepth without scene type supervision. ZoeDepth performance degrades significantly when the scene type (indoor or outdoor) is not provided.
\ours~learns well without such supervision.}
\label{tab:ablation-router}
\end{table}

\vspace*{0.2cm}
\noindent
\textbf{No FOV augmentation or conditioning.}
ZoeDepth found that without scene-type supervision for the experts (i.e., \textit{Auto Router}), ZoeDepth's performance degrades, even for in-domain data. To compare against ZoeDepth in this setting, we fine-tune a model on NYU and KITTI without FOV augmentations or conditioning.
Interestingly, \ours~performs relatively well in this setting for in-domain data (Table \ref{tab:ablation-router}). Nevertheless, as shown in Table \ref{tab:ablation-fov-aug-cond},
OOD performance is better with FOV augmentation and conditioning.

\vspace*{0.2cm}
\noindent
\textbf{$\veps$ vs $\vv$ diffusion parameterization.} Inference latency is a concern with diffusion models for vision. DDVM \cite{ddvm}, for example, uses 128 denoising steps for depth estimation which can be prohibitive.
We find that using the $v$-parameterization dramatically reduces the number of denoising steps required for good performance. 
As shown in Table \ref{tab:ablation-eps-v}, $\veps$-parameterization requires 64 denoising steps to match the performance of a model with $\vv$-parameterization using only 1 denoising step. 
Intuitively, $\vv$-parameterization ensures that the model accurately recovers the signal at both ends of the noise schedule, unlike $\veps$-parameterization where estimating the noise is easy for low SNR inputs.

\begin{table}[t]
\footnotesize
\setlength{\tabcolsep}{3pt} 
\centering
\begin{tabular}{c|cc|cc}
\toprule
& \multicolumn{2}{c|}{NYU} & \multicolumn{2}{c}{KITTI} \\
Num denoising steps & $\veps $   & $\vv$ & $\veps$   & $\vv$ \\ \midrule

1 & 2.374 & 0.077 & 0.596 & 0.056 \\
4 & 1.484 & 0.075 & 0.406 & 0.055 \\
16 & 0.409 & 0.074 & 0.141 & 0.055 \\
64 & 0.077 & 0.074 & 0.056 & 0.055 \\
\bottomrule
\end{tabular}
\caption{Comparison of relative error on NYU and KITTI for models trained with $\veps$- and $\vv$- parameterization. Both are fine-tuned on NYU and KITTI  without FOV augmentation or conditioning.}
\label{tab:ablation-eps-v}
\end{table}
\section{Conclusion}
We propose a generic diffusion-based monocular metric depth generator with no task-specific inductive biases and no specialized architectures for handling diverse indoor and outdoor scenes. Our log-scale depth parameterization adequately allocates representation capacity to different depth ranges. We advocate augmenting the FOV of training data through simple cropping/uncropping to enable generalization to fields-of-view beyond those in the training datasets and show that simply uncropping with noise padding is effective for simulating a larger FOV. We find that conditioning on the FOV is essential for disambiguating depth-scale. We further propose a new fine-tuning dataset mixture that dramatically improves performance. With these innovations combined, we establish a new state-of-the-art outperforming the prior work of 
ZoeDepth across all zero-shot and in-domain datasets by a substantial margin.

\subsection*{Acknowledgements} 
We thank Jon Shlens, Kevin Swersky, Shekoofeh Azizi and Cristina Vasconcelos for their detailed feedback and help with figures. We  also thank Ting Chen, Daniel Watson, Forrester Cole and others in Google DeepMind and Google Research for helpful discussions.\\ 

{
    \small
    \bibliographystyle{ieeenat_fullname}
    \bibliography{main}
}

\appendix

\vspace{10cm}

\begin{figure*}[t]
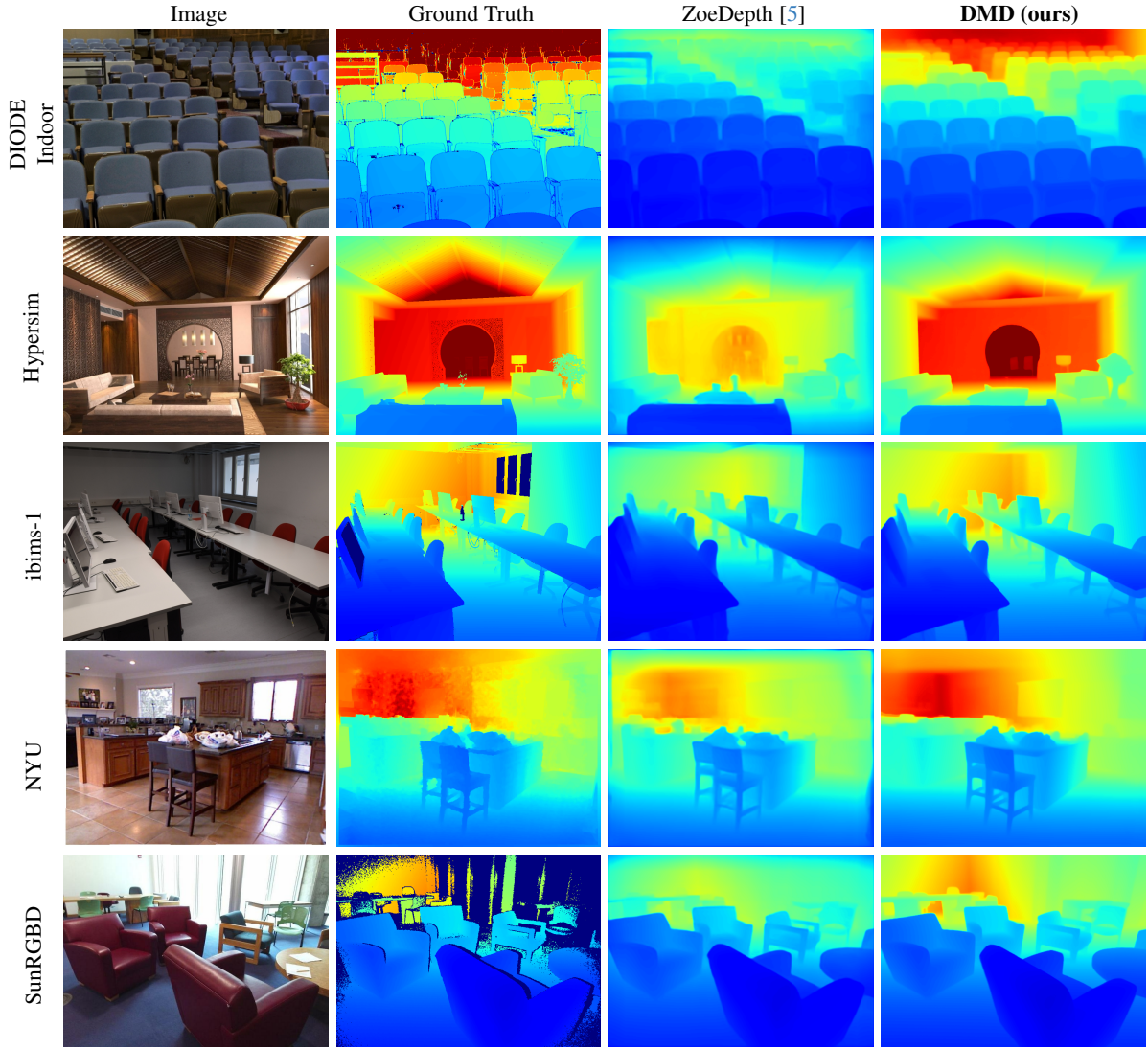

\setlength\tabcolsep{0.5pt}
\center
\small

\newcommand{\imgwidth}{0.21\textwidth}
\newcommand{\image}[3]{
\includegraphics[width=\imgwidth]{figures/qualitative_comparison/images/#1/#2_#3.png}
}
\newcommand{\imagejpg}[3]{
\includegraphics[width=\imgwidth]{figures/qualitative_comparison/images/#1/#2_#3.jpg}
}

\newcommand{\row}[2]{
\image{#1}{#2}{image} & \image{#1}{#2}{gt} & \image{#1}{#2}{depth_zd} & \image{#1}{#2}{depth_ours}
}
\newcommand{\rowrgbjpg}[2]{
\imagejpg{#1}{#2}{image} & \image{#1}{#2}{gt} & \image{#1}{#2}{depth_zd} & \image{#1}{#2}{depth_ours}
}

\begin{tabular}{ccccc}
    & Image & Ground Truth & ZoeDepth \cite{bhat2023zoedepth} & \textbf{\ours~(ours)} \\
    \rotatebox{90}{\hspace{2.3em} \makecell{DIODE\\Indoor}} &
    \rowrgbjpg{diode_indoor}{00187} \\
    \rotatebox{90}{\hspace{2em} Hypersim} &
    \rowrgbjpg{hypersim}{03499} \\
    \rotatebox{90}{\hspace{2.4em} ibims-1} & 
    \rowrgbjpg{ibims}{00027} \\
    \rotatebox{90}{\hspace{2.2em} NYU} & 
    \rowrgbjpg{nyu}{00234} \\
    \rotatebox{90}{\hspace{1.6em} SunRGBD} & 
    \rowrgbjpg{sunrgbd}{01352} \\
\end{tabular}
\vspace*{-0.2cm}
\caption{Qualitative comparison between our method and ZoeDepth \cite{bhat2023zoedepth} on indoor scenes.
Compared with ZoeDepth, our method estimates depths at more accurate scale over diverse datasets.}
\label{fig:ours_zd_samples_indoor2}
\end{figure*}

\section{Additional qualitative examples}

\begin{figure*}[t]
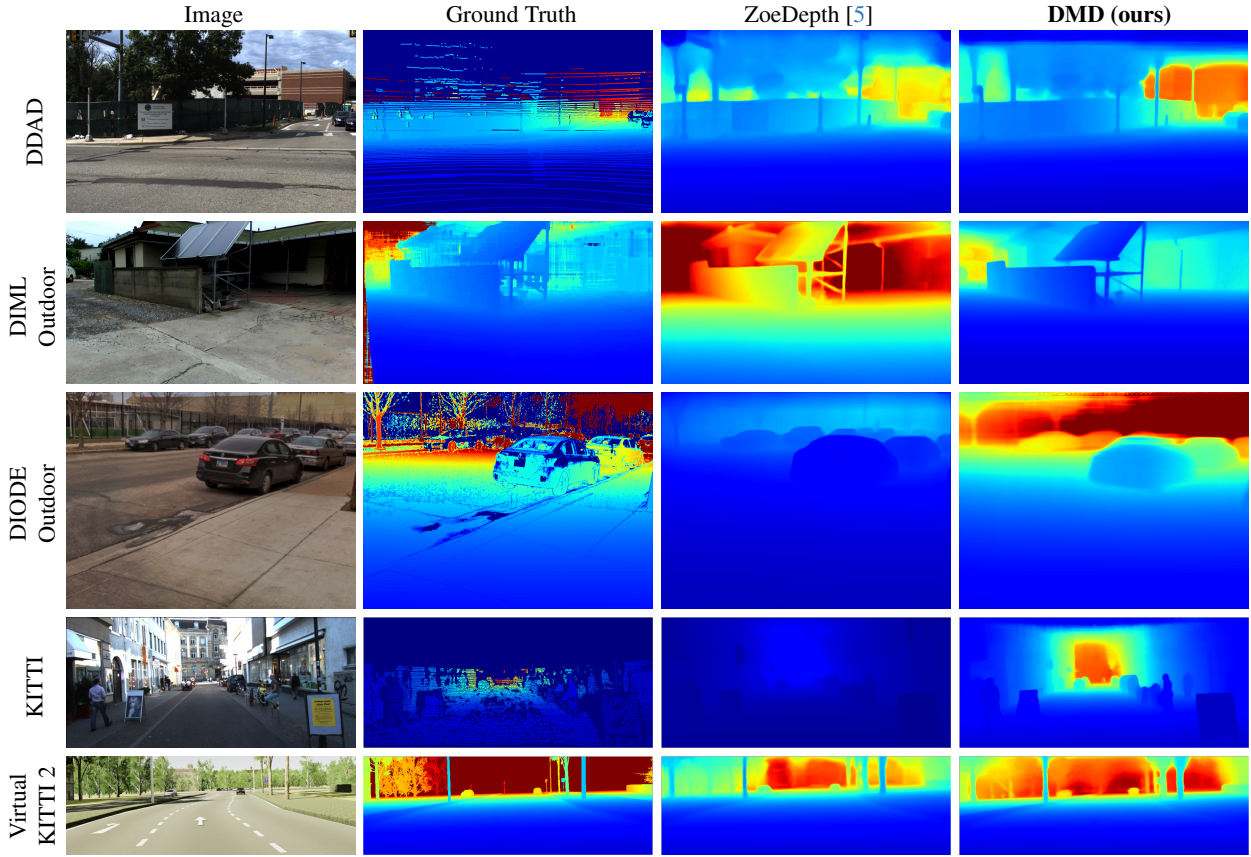

\setlength\tabcolsep{0.5pt}
\center
\small

\newcommand{\imgwidth}{0.22\textwidth}

\newcommand{\image}[3]{
\includegraphics[width=\imgwidth]{figures/qualitative_comparison/images/#1/#2_#3.png}
}

\newcommand{\imagejpg}[3]{
\includegraphics[width=\imgwidth]{figures/qualitative_comparison/images/#1/#2_#3.jpg}
}

\newcommand{\row}[2]{
\image{#1}{#2}{image} & \image{#1}{#2}{gt} & \image{#1}{#2}{depth_zd} & \image{#1}{#2}{depth_ours}
}

\newcommand{\rowrgbjpg}[2]{
\imagejpg{#1}{#2}{image} & \image{#1}{#2}{gt} & \image{#1}{#2}{depth_zd} & \image{#1}{#2}{depth_ours}
}

\begin{tabular}{ccccc}
    & Image & Ground Truth & ZoeDepth \cite{bhat2023zoedepth} & \textbf{\ours~(ours)} \\
    \rotatebox{90}{\hspace{1.7em} DDAD} &
    \rowrgbjpg{ddad}{00900} \\
    \rotatebox{90}{\hspace{1.3em} \makecell{DIML \\ Outdoor}} &
    \rowrgbjpg{diml_outdoor}{00424} \\
    \rotatebox{90}{\hspace{2.5em} \makecell{DIODE \\ Outdoor}} & 
    \rowrgbjpg{diode_outdoor}{00346} \\
    \rotatebox{90}{\hspace{1em} KITTI} & 
    \rowrgbjpg{kitti}{00529} \\
    \rotatebox{90}{\hspace{0.1em}  \makecell{Virtual \\ KITTI 2}} & 
    \rowrgbjpg{vkitti2}{00371} \\
\end{tabular}

\vspace*{-0.2cm}
\caption{Qualitative comparison between \ours~and ZoeDepth \cite{bhat2023zoedepth} on outdoor scenes. Compared with ZoeDepth \cite{bhat2023zoedepth}, our method is able to estimate a more accurate depth scale.}

\label{fig:ours_zd_samples_outdoor2}
\end{figure*}

In \cref{fig:ours_zd_samples_indoor2} and \cref{fig:ours_zd_samples_outdoor2}, we provide additional qualitative  comparison between DMD and ZoeDepth on indoor and outdoor datasets respectively. Our method consistently captures more accurate metric-scale depth in various indoor and outdoor scenes.

\begin{table*}[t]
\footnotesize
\centering
\setlength{\tabcolsep}{2.5pt} 
\begin{tabular}{lccccc|ccccc}
\toprule
& \multicolumn{5}{c|}{\textbf{Indoors}} & \multicolumn{5}{c}{\textbf{Outdoors}} \\
Experiment & NYU & SunRGBD & DIODE Indoor & ibims-1 & Hypersim & KITTI & DIML Outdoor & DIODE Outdoor & Virtual KITTI 2 & DDAD \\
\midrule
\oursnk~true FOV & 0.076 & 0.109 & 0.298 & 0.130 & 0.382 & 0.055 & 0.300 & 0.628 & 0.093 & 0.122 \\
\oursnk~est. FOV & 0.077 & 0.124 & 0.302 & 0.140 & 0.407 & 0.062 & 1.490 & 0.639 & 0.114 & 0.132 \\
\midrule
\oursmix~true FOV & 0.072 & 0.091 & 0.291 & 0.118 & 0.318 & 0.053 & 0.190 & 0.553 & 0.092 & 0.108 \\
\oursmix~est. FOV & {0.072} & 0.136 & {0.291} & {0.114} & {0.361} & 0.061 & 1.049 & {0.560} & 0.131 & {0.119} \\
\midrule
FOV est. error (degrees) & 1.520 & 5.114 & 0.735 & 2.339 & 4.224 & 0.842 & 27.998 & 3.459 & 4.142 & 4.384 \\
\bottomrule
\end{tabular}
\vspace{-6pt}
\caption{Comparison of relative error (REL) using an estimated (est.) FOV versus the true FOV. Even with a simple FOV estimator the depth estimation results using an estimated FOV are competitive to those with the true FOV, except for DIML Outdoor where our FOV estimator struggles to predict correct FOV as shown by the large error in FOV estimation.}
\label{tab:estimated_fov}
\end{table*}

\section{Handling unknown field-of-view}
While RGB camera intrinsics are available for most practical uses of monocular depth estimators (e.g. cell phones, robot platforms or self-driving cars), they may sometimes be unknown (e.g. internet images or generative imagery). One solution to handle the unknown FOV would be to estimate the camera intrinsics from the RGB image.

To test this, we train a simple model for predicting the vertical field-of-view (FOV). The model consists of the encoder of a pre-trained Palette \cite{sahariac-palette} model followed by a spatial average pooling layer and a linear head that predicts a scalar. It is trained to regress to $\tan(\theta/2)$, where $\theta$ is the vertical FOV, using a $L_1$ loss. We train on a mix of the NYU and KITTI datasets employing the same FOV augmentation strategy as used for \oursnk.

Table \ref{tab:estimated_fov} compares the depth estimation performance of our models when using the estimated FOV versus the true FOV. Despite the simplistic design of our FOV estimator, the depth estimation performance when using the estimated FOV is competitive to that when using the true FOV, except for the DIML outdoor dataset, where the large error in FOV estimation leads to a much worse relative error for depth.
This might be due to the significantly larger FOV of DIML outdoor, compared to other outdoor datasets, which could be challenging for our simple FOV regressor to generalize to.
We hypothesize that incorporating more sophisticated camera intrinsic prediction models, such as those proposed by \cite{Hold_Geoffroy_2018,perspectivefields2023}, could lead to further improvements in FOV estimation accuracy, and thereby better metric depth estimates. However, we defer a thorough investigation of this approach to future work.

\section{Additional training details}
We perform the first stage of supervised training for 1.5M steps with a learning rate of $1\times10^{-4}$ and the second (final) stage for 50k steps with a learning rate of $3\times10^{-5}$. For the FOV augmentation in the second stage we randomly uniformly sample a scale in [0.8, 1.5] to crop/uncrop the image and depth maps. 

\end{document}